\documentclass[letterpaper]{article} 
\usepackage{aaai24}  
\usepackage{times}  
\usepackage{helvet}  
\usepackage{courier}  
\usepackage[hyphens]{url}  
\usepackage{graphicx} 
\usepackage{natbib}  
\usepackage{caption} 
\usepackage{algorithm}
\usepackage{algorithmic}

\usepackage{newfloat}
\usepackage{listings}
\DeclareCaptionStyle{ruled}{labelfont=normalfont,labelsep=colon,strut=off} 
\floatstyle{ruled}
\newfloat{listing}{tb}{lst}{}
\floatname{listing}{Listing}

\title{Surrogate Modelling for Sea Ice Concentration using Lightweight Neural Ensemble}

\author {
    Julia Borisova\textsuperscript{\rm 1},
    Nikolay O. Nikitin\textsuperscript{\rm 1}
}
\affiliations {
    \textsuperscript{\rm 1}ITMO, 49 Kronverksky Pr. St. Petersburg, 197101, Russian Federation\\
    jul.borisova@itmo.ru, nnikitin@itmo.ru
}

\begin{document}
\maketitle
\begin{abstract}

The modeling and forecasting of sea ice conditions in the Arctic region are important tasks for ship routing, offshore oil production, and environmental monitoring. We propose the adaptive surrogate modeling approach named \textit{LANE-SI} (\textbf{L}ightweight \textbf{A}utomated \textbf{N}eural \textbf{E}nsembling for \textbf{S}ea \textbf{I}ce) that uses ensemble of relatively simple deep learning models with different loss functions for forecasting of spatial distribution for sea ice concentration in the specified water area. 

Experimental studies confirm the quality of a long-term forecast based on a deep learning model fitted to the specific water area is comparable to resource-intensive physical modeling, and for some periods of the year, it is superior. We achieved a 20\% improvement against the state-of-the-art physics-based forecast system SEAS5 for the Kara Sea.

\end{abstract}

\section{Introduction}

In recent years, Arctic modeling has become a focus of increased interest for researchers in various scientific fields. The sea ice melting process, in conjunction with global warming, determines the ecological state of the region, including biodiversity conservation. Ice concentration distribution plays a key role in laying tracks by icebreakers and advancing cargo and research ships.

Predictive modeling of ice conditions in the Arctic is considered in different time scales. Short-term forecasts involve the assimilation of operational information and reproducing accurate data for decision-making. This type of forecasting is limited to intervals from several hours to several months. Long-term forecasting is carried out one or several seasons in advance to enable advance planning, which is essential in the industrial sector \cite{harsem2015oil}. In predicting the timing of opening and freezing of the water area, statistical modeling is carried out based on retrospective observational data \cite{wu2023statistical}.

Physical modeling became the classical approach to the simulation of natural environments because of advantages on the spatial and temporal scales and the ability to predict and analyze different scenarios \cite{chen2023sea}. It utilizes complex numerical methods to solve the equations of motion, thermodynamics, and radiation \cite{stroeve2007arctic}. Despite the robustness and explainability of physical modeling, its application is often complicated by high computational cost, the need for boundary and initial conditions for many variables, and complex parameterization for a specific territory. Years of active development of earth remote sensing have contributed to the accumulation of spatiotemporal datasets for environmental systems, which can be used to train various data-driven models, including deep neural networks. While deep learning models can provide high-quality environmental forecasts \cite{lam2023learning}, the computational cost for models' training to satisfy the task-specific requirements (water area, time and spatial resolution, forecast length, objective function) can be excessive. For this reason, we aimed to provide a more lightweight surrogate alternative to existing solutions for sea ice forecasting.

This paper presents a deep learning approach \textit{LANE-SI} that uses an ensemble of convolutional neural networks (CNN) for long-term predictive modeling of sea ice concentration. Satellite data with post-correction - OSI SAF Global Sea Ice Concentration (SSMIS) product \cite{tonboe2016product} was taken as training data. Results are presented for the Kara Sea and part of the Barents Sea region of the Arctic. The developed model is lightweight due to its simple architecture and is also undemanding in input data - only the pre-history of the target parameter is required. A comparison was made with the ECMWF's fifth-generation physics-based seasonal forecast system - SEAS5 \cite{johnson2019seas5}. The comparison results demonstrate that the surrogate model reproduces the absolute values of ice concentration and their spatial distribution with a quality comparable to the SEAS5 forecast. The surrogate model reproduces the position of the ice edge better than the SEAS5 forecast, both on average contour points position and the position of each point separately. 

 The paper is structured as follows: Section 2 reviews the key related work, Section 3 describes the proposed method for the design of surrogate sea ice forecasting models named \textit{LANE-SI}, Section 4 contains the experimental results, and Section 5 presents concluding remarks for the proposed methods.

\section{Related Work}

The classic method for modeling sea ice parameters is numerical modeling based on systems of differential equations. Over the years of technological development, a large number of both specific solutions that reproduce local ice dynamics \cite{girard2011new} and general methods of numerical hydrometeorological forecasting \cite{wang2013seasonal, zhang2003modeling} have been developed. State-of-the-art physical models like NEMO-SEAICE \cite{madec2017nemo} make it possible to reproduce ice conditions both at the global \cite{hvatov2019adaptation} and regional \cite{pemberton2017sea} levels. Considering the universality of numerical models, it should be noted that their adaptation to a specific territory requires significant effort from a subject specialist and considerable computing resources, as well as setting boundary and initial conditions, which raises the need for high-quality initial data, which is not always available for the water area interest.

There has been a rapid development of data-driven approaches in recent years. Model complexity varies from linear relationship searches \cite{kapsch2014importance} and regression models \cite{ahn2014statistical} to deep learning networks \cite{chi2017prediction}. Regression models are limited in their applicability due to the enormously increasing complexity when moving to grid calculations. Using neural networks based on simple convolutional \cite{kim2020prediction} and more complex U-Net \cite{ali2022mt} architectures provide high-resolution output images and allow a more detailed forecast. However, most high-quality existing solutions require a large amount of additional input data and computing power \cite{andersson2021seasonal, ali2022mt} for training and network prediction, significantly limiting their applicability for specific water areas.

 It is known that combining a set of simple models into ensembles makes it possible to achieve higher quality forecasting, including in the subject area of hydrometeorology \cite{ali2021sea, kim2018satellite}. We assume that ensembling of simple neural networks can lead us to better quality at lower computational costs \cite{zhou2002ensembling}. Thus, the question arises of finding a balance between the complexity of the model and the quality of the forecast.

\section{LANE-SI}

To achieve a better quality of high-resolution long-term sea ice forecasting in a computationally cheap way, we developed the \textit{LANE-SI} approach that allows designing the predictive model for sea ice in specific water areas. We used the convolutional neural network for a non-standard task - time-spatial forecasting. In contrast to LSTM architectures, which allow the extraction of features only from time series \cite{choi2019artificial}, convolution emphasizes the spatial distribution of the parameter and reduces the risk of stagnation over time. The surrogate model was implemented to carry out long-term forecasting with a forecast horizon of one year. We aimed to make the model as flexible as possible to adapt to specific water areas to reduce the computational cost against neural surrogate models that cover the entire Arctic region \cite{andersson2021seasonal}.


In LANE-SI, the dataset for training a surrogate data-driven predictive model for ice concentration is prepared similarly to the lagged transformation applied to time series. Previous $k$ steps are used to predict $n$ steps ahead using the pre-history of sea ice. Each time step is characterized by an image describing the concentration in each cell. 

The surrogate model is represented by an ensemble that aggregates two CNNs trained with different loss functions and an inertial forecast - concentration values averaged for each day of the year over the previous five years.

\subsection{CNN for sea ice forecasting}

\begin{figure}[!h]
    \centering{\includegraphics[scale=0.8]{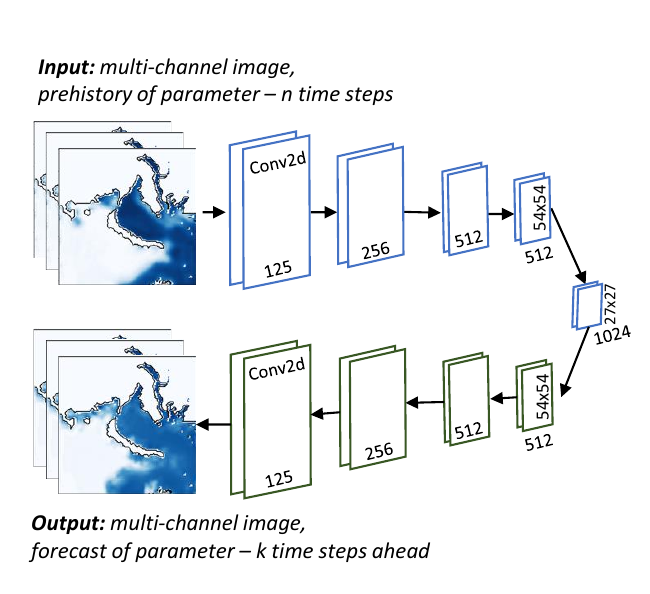}}
    \caption{Architecture of deep model for sea ice concentration forecasting. The specific values of layer shape are described for the experimental domain (125x125) used in Section~\ref{sec_concl}.}
    \label{fig:schema_CNN}
\end{figure}

The basic structure of the deep neural networks used in \textit{LANE-SI} include five convolution layers with ReLU activation functions \cite{agarap2018deep}. Postprocessing with cutting to the range was used due to the inconsistency of using the sigmoid activation function to ensure that the output image values range from 0 to 1 \cite{wang2017sea}. The structural scheme of CNN is presented in Figure~\ref{fig:schema_CNN}. The specific size of layers is configured in an automated way.

In the modeling of spatiotemporal data, it is important not only to reproduce the absolute values but also to measure the proximity of the spatial distribution of values to the distribution in the target image. Therefore, CNNs were trained on two loss functions:
\begin{itemize}
\item \textbf{Mean Absolute Error (MAE, L1Loss)} - represents the closeness of the absolute values of each pixel of each prediction element with the target image;
\item \textbf{Structural Similarity Index (SSIM)} - represents the similarity local patterns of pixel intensities \cite{wang2004image} and shows similarity in the spatial distribution of the parameter at predicted and target images. 
\end{itemize}

The CNN accepts multi-channel images with several years of ice concentration pre-history.
At the output, the CNN provides images of the same spatial resolution as the input ones, but the number of channels is 52. It corresponds to a one-year forecast with a time sampling of seven days.

\subsection{Neural ensembling of forecasts}

The multi-model ensemble was used to improve the quality of the surrogate model in \textit{LANE-SI}. The first CNN reflects the spatial distribution of the parameter (SSIM), and the second reproduces the parameter's absolute values (MAE). Also, a naive forecast in the form of repeating the average values for five years for each day of the year was added to reproduce long-term dynamics not represented in pre-history.

Three types of ensembling methods were tested by quality comparison with the same metrics (MAE, SSIM):
\begin{itemize}
\item Simple weighted ensemble, based on linear regression: each pixel in a predicted multi-channel image is considered an independent variable. Predicted values for each pixel of three separated models are summed with coefficients that provide less error on the delayed part of train data;
\item CNN trained with MAE loss function uniting the elements of the ensemble in the form of multi-channel images into one prediction;
\item CNN trained with SSIM loss function unites the ensemble's elements in the form of multi-channel images into one prediction.
\end{itemize}

\begin{figure*}[!h]
    \centering{\includegraphics[scale=0.9]{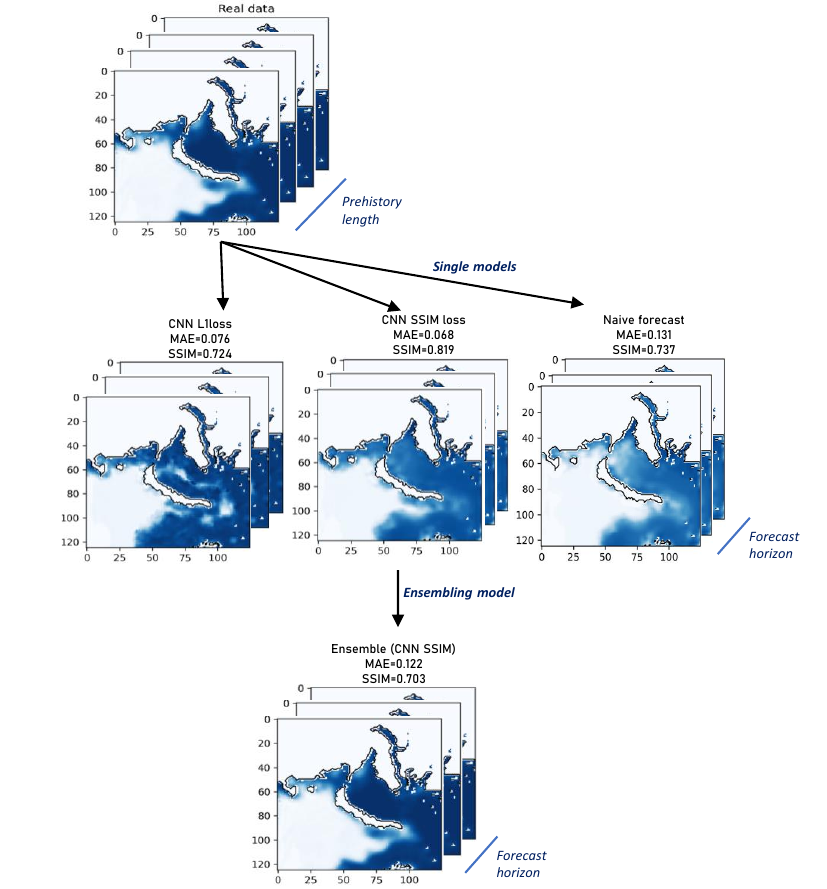}}
    \caption{The structure of the proposed ensemble model (first layer - predictive model, second layer - neural ensembling model).}
    \label{fig:schema_ens_struct}
\end{figure*}

The general scheme for an approach that combines single models into using neural ensembling is presented in Figure~\ref{fig:schema_ens_struct}.

\section{Experiments and Results}

\subsection{Dataset}

\begin{figure}[!h]
    \centering{\includegraphics[scale=0.35]{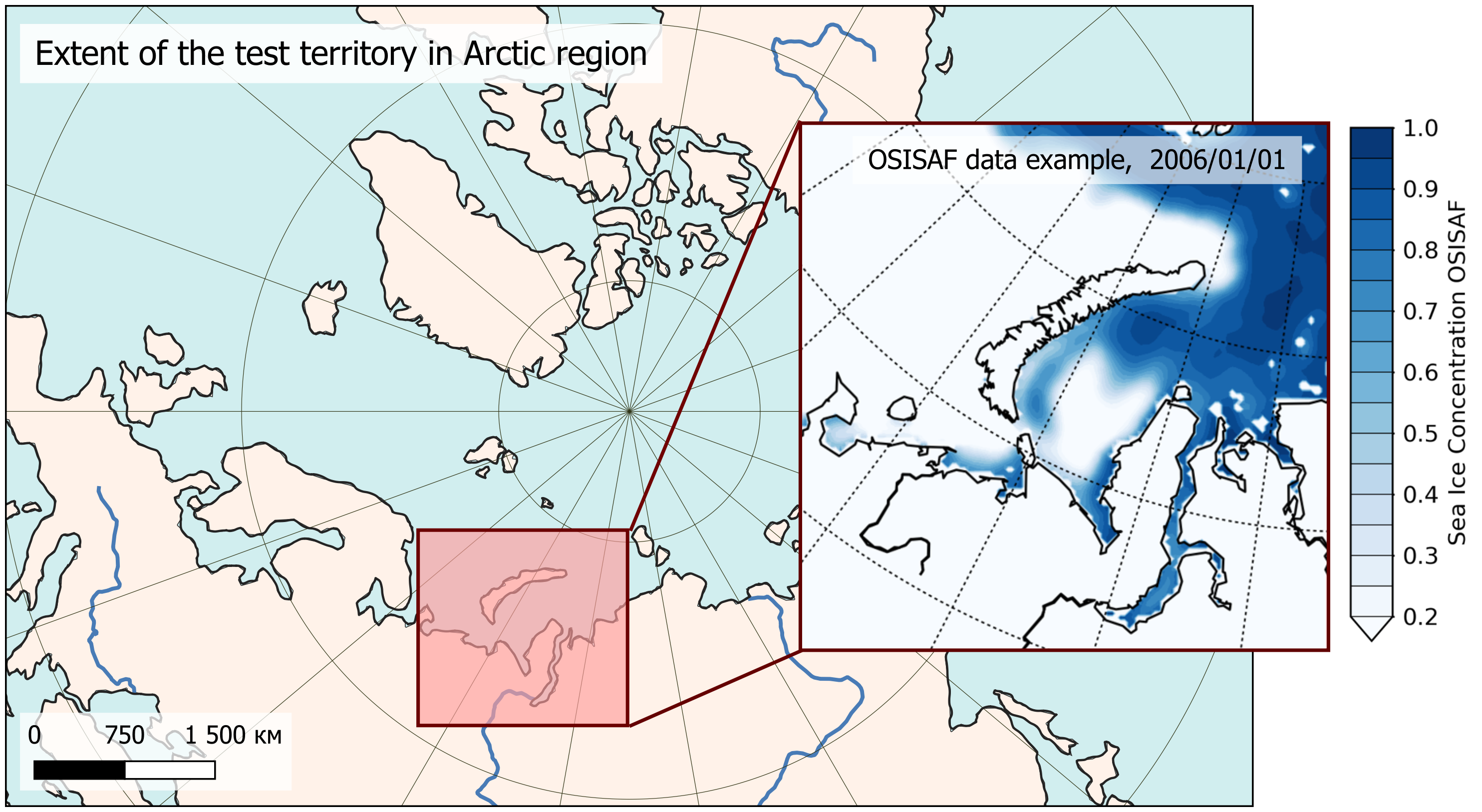}}
    \caption{Spatial extent of testing domain with OSI SAF ice concentration data sample on 2006/01/01.}
    \label{fig:schema_arct}
\end{figure}

We use OSI SAF Global Sea Ice Concentration (SSMIS) product as training data for sea ice concentration forecasting in specific water areas. The time resolution of the product is one day; however, for the problem of long-term forecasting, such discretization is redundant, and therefore the time resolution was sparsed to weekly. Data were normalized to a regular grid with a spatial resolution of 14 km using bilinear interpolation. An area with the Kara Sea and part of the Barents Sea was chosen as a test domain (as presented in Figure~\ref{fig:schema_arct}).

We conduct the experimental evaluation of \textit{LANE-SI} for the Kara Sea using SSMIS for 1996 to 2022 years. Primary forecasting models were trained on the data time range from 1996 to 2009. The ensembling model was trained on predictions of single models from 2010 to 2015. They were additionally re-trained on the entire training set from 1996 to 2015 to obtain better forecasts using single models. Validation was made as an out-of-sample one-year-ahead forecast for 2006-2022 compared with naive forecast and actual data.

The training time for \textit{LANI-SI} in the described setting on a local machine with an NVIDIA GeForce RTX 4080 video card and an AMD Ryzen 9 5950X processor (16 x 3.4 GHz) was approximately 450 minutes.

\subsection{LANE-SI against separate data-driven models}

The quality of the surrogate model (design with \textit{LANE-SI}) for the Kara Sea was analyzed compared to the naive forecast for both the final ensemble model and individual ensemble components. Table~\ref{tab_comp} presents numerical estimates of metrics averaged over the years. 

\begin{table*}[t]
\centering
\begin{tabular}{|c|cccc|cccc|}
\hline
Metric                      & \multicolumn{4}{c|}{Mean Absolute Error (MAE)}            & \multicolumn{4}{c|}{Structural Similarity Index (SSIM)}                         \\ \hline
 
 Year & \multicolumn{1}{c|}{ Surrogate} & \multicolumn{1}{c|}{ \begin{tabular}[c]{@{}c@{}}CNN \\ L1Loss\end{tabular}} & \multicolumn{1}{c|}{ \begin{tabular}[c]{@{}c@{}}CNN \\ SSIM loss\end{tabular}} &  \begin{tabular}[c]{@{}c@{}}Naive \\ forecast\end{tabular} & \multicolumn{1}{c|}{Surrogate} & \multicolumn{1}{c|}{\begin{tabular}[c]{@{}c@{}}CNN \\ L1Loss\end{tabular}} & \multicolumn{1}{c|}{\begin{tabular}[c]{@{}c@{}}CNN \\ SSIM loss\end{tabular}} & \begin{tabular}[c]{@{}c@{}}Naive \\ forecast\end{tabular} \\ \hline

 2016 & \multicolumn{1}{c|}{ 0,064}     & \multicolumn{1}{c|}{\bfseries 0,044}                                                 & \multicolumn{1}{c|}{ 0,048}                                                    &  0,086                                                     & \multicolumn{1}{c|}{\bfseries 0,807}     & \multicolumn{1}{c|}{0,726}                                                 & \multicolumn{1}{c|}{0,796}                                                    & 0,723                                                     \\ \hline

 2017 & \multicolumn{1}{c|}{\bfseries 0,056}     & \multicolumn{1}{c|}{ 0,059}                                                 & \multicolumn{1}{c|}{\bfseries 0,056}                                                    &  0,064                                                     & \multicolumn{1}{c|}{\bfseries0,810}     & \multicolumn{1}{c|}{0,701}                                                 & \multicolumn{1}{c|}{0,784}                                                    & 0,758                                                     \\ \hline

 2018 & \multicolumn{1}{c|}{\bfseries 0,052}     & \multicolumn{1}{c|}{ 0,061}                                                 & \multicolumn{1}{c|}{\bfseries 0,052}                                                    &  0,06                                                      & \multicolumn{1}{c|}{\bfseries 0,808}     & \multicolumn{1}{c|}{0,690}                                                 & \multicolumn{1}{c|}{0,776}                                                    & 0,768                                                     \\ \hline

 2019 & \multicolumn{1}{c|}{\bfseries 0,045}     & \multicolumn{1}{c|}{ 0,061}                                                 & \multicolumn{1}{c|}{ 0,056}                                                    &  0,055                                                     & \multicolumn{1}{c|}{\bfseries 0,823}     & \multicolumn{1}{c|}{0,689}                                                 & \multicolumn{1}{c|}{0,782}                                                    & 0,780                                                     \\ \hline

 2020 & \multicolumn{1}{c|}{ 0,085}     & \multicolumn{1}{c|}{\bfseries 0,064}                                                 & \multicolumn{1}{c|}{ 0,079}                                                    &  0,104                                                     & \multicolumn{1}{c|}{0,509}     & \multicolumn{1}{c|}{\bfseries 0,595}                                                 & \multicolumn{1}{c|}{0,522}                                                    & 0,469                                                     \\ \hline

 2021 & \multicolumn{1}{c|}{\bfseries 0,07}      & \multicolumn{1}{c|}{ 0,089}                                                 & \multicolumn{1}{c|}{ 0,079}                                                    &  0,087                                                     & \multicolumn{1}{c|}{\bfseries0,526}     & \multicolumn{1}{c|}{0,525}                                                 & \multicolumn{1}{c|}{0,500}                                                    & 0,496                                                     \\ \hline

 2022 & \multicolumn{1}{c|}{\bfseries 0,064}     & \multicolumn{1}{c|}{ 0,074}                                                 & \multicolumn{1}{c|}{ 0,082}                                                    &  0,085                                                     & \multicolumn{1}{c|}{0,532}     & \multicolumn{1}{c|}{\bfseries 0,552}                                                 & \multicolumn{1}{c|}{0,492}                                                    & 0,494                                                     \\ \hline
                                                  
\end{tabular}
\caption {Comparison of surrogate ensemble model and single CNN models by MAE and SSIM metric mean for each year}
\label{tab_comp}
\end{table*}

As the table shows, the surrogate ensemble model achieves a higher forecast quality than single models according to the SSIM metric for all years and almost all years according to the MAE metric. 

\begin{figure*}[!h]
    \centering{\includegraphics[scale=0.8]{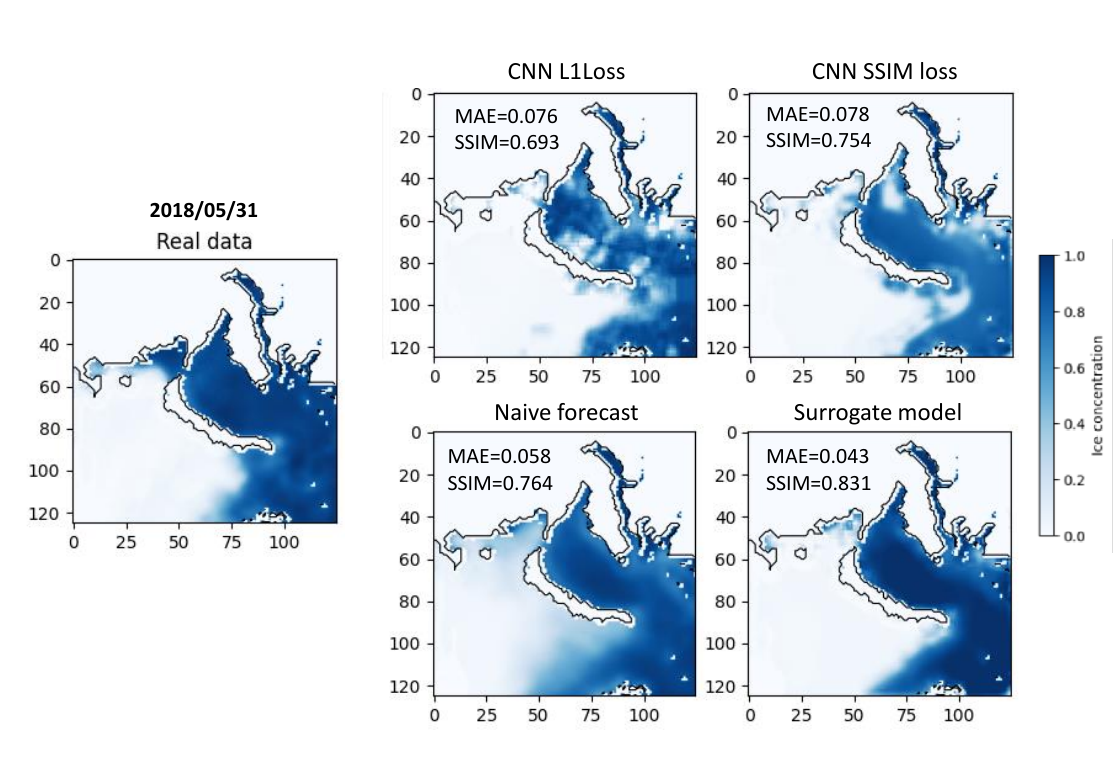}}
    \caption{Comparison of the spatial distribution of forecasted ice concentration by ensemble surrogate model and separate models with real data for 31.05.2018}
    \label{fig:schema_ens}
\end{figure*}

Figure~\ref{fig:schema_ens} presents an example of a spatial comparison of different data-driven forecasts (\textit{LANE-SI}-designed ensemble and separate neural models) with actual data for a single forecast step. 

This example shows the shortcomings of single models. The CNN with the L1 loss function reproduces the ice field with large granularity despite its good fit into the absolute values on average over the image. CNN with SSIM loss function and naive forecast tends to over-smooth the image and underestimate the absolute values of the parameter. Thus, the ensemble model combines single models while avoiding their inherent disadvantages.

\subsection{LANE-SI against physics-based forecast}

To objectively assess the quality of the surrogate model, it was compared with the state-of-the-art (SOTA) model based on differentiation equations - the predictive model SEAS5. 

The maximum forecast horizon of the SEAS5 model is nine months (against 12 months for the design surrogate model). Therefore, we configured both forecasts to start on January 1 of each year. The error was calculated on average for each season (three months) out of three fully predicted. 

Table~\ref{tab_res} presents the comparison results for the entire SEAS5's available archives (2020-2022).

\begin{table}[]
\begin{tabular}{|c|cc|cc|}
\hline
Metric                                                     & \multicolumn{2}{c|}{\begin{tabular}[c]{@{}c@{}}Mean Absolute \\ Error (MAE)\end{tabular}} & \multicolumn{2}{c|}{\begin{tabular}[c]{@{}c@{}}Structural Similarity \\ Index (SSIM)\end{tabular}} \\ \hline
\begin{tabular}[c]{@{}c@{}}Year \\ and quater\end{tabular} & \multicolumn{1}{c|}{SEAS5}  & \begin{tabular}[c]{@{}c@{}}Surrogate \\ model\end{tabular}  & \multicolumn{1}{c|}{SEAS5}       & \begin{tabular}[c]{@{}c@{}}Surrogate \\ model\end{tabular}      \\ \hline
2020Q1                                                     & \multicolumn{1}{c|}{\bfseries 0,097}  & 0,103                                                       & \multicolumn{1}{c|}{\bfseries 0,602}       & 0,516                                                           \\ \hline
2020Q2                                                     & \multicolumn{1}{c|}{0,113}  & \bfseries 0,108                                                       & \multicolumn{1}{c|}{\bfseries 0,542}       & 0,474                                                           \\ \hline
2020Q3                                                     & \multicolumn{1}{c|}{0,082}  & \bfseries 0,021                                                       & \multicolumn{1}{c|}{\bfseries 0,583}       & 0,571                                                           \\ \hline
2021Q1                                                     & \multicolumn{1}{c|}{\bfseries 0,099}  & 0,103                                                       & \multicolumn{1}{c|}{\bfseries 0,630}       & 0,521                                                           \\ \hline
2021Q2                                                     & \multicolumn{1}{c|}{\bfseries 0,078}  & 0,094                                                       & \multicolumn{1}{c|}{\bfseries 0,610}       & 0,490                                                           \\ \hline
2021Q3                                                     & \multicolumn{1}{c|}{0,063}  & \bfseries 0,029                                                       & \multicolumn{1}{c|}{\bfseries 0,626}       & 0,560                                                           \\ \hline
2022Q1                                                     & \multicolumn{1}{c|}{0,103}  & \bfseries 0,092                                                       & \multicolumn{1}{c|}{\bfseries 0,587}       & 0,540                                                           \\ \hline
2022Q2                                                     & \multicolumn{1}{c|}{0,095}  & \bfseries 0,081                                                       & \multicolumn{1}{c|}{\bfseries 0,546}       & 0,508                                                           \\ \hline
2022Q3                                                     & \multicolumn{1}{c|}{0,074}  & \bfseries 0,025                                                       & \multicolumn{1}{c|}{\bfseries 0,576}       & 0,563                                                           \\ \hline
\end{tabular}
\caption {Comparison of surrogate ensemble model and SEAS5 forecast for first three quarters of each year}
\label{tab_res}
\end{table}

As the table shows, forecasting using a surrogate model designed by \textit{LANE-SI} is comparable in quality to the physics-based forecast and, in some cases, can even surpass it due to the better adaptation to specific water areas.

In addition to the metrics used for training (SSIM, MAE), a metric for comparing the similarity of the ice edge was used to assess the quality. The ice edge is the contour of the spatial position of ice with a concentration of more than 0.8. An expert can choose the threshold for switching to binary data (it depends on the scenario of the forecast usage). In this case, the ice edge's position is represented by a set of points with coordinates of a fixed length. The distance metric is calculated as a mean of the signed distance between each point of one contour and the nearest contour edge of the second contour \cite{brahmbhatt2013practical}. This metric was introduced because it is more indicative for assessing the spatial distribution of ice concentration values critical for navigation.

\begin{figure*}[!h]
    \centering{\includegraphics[scale=0.9]{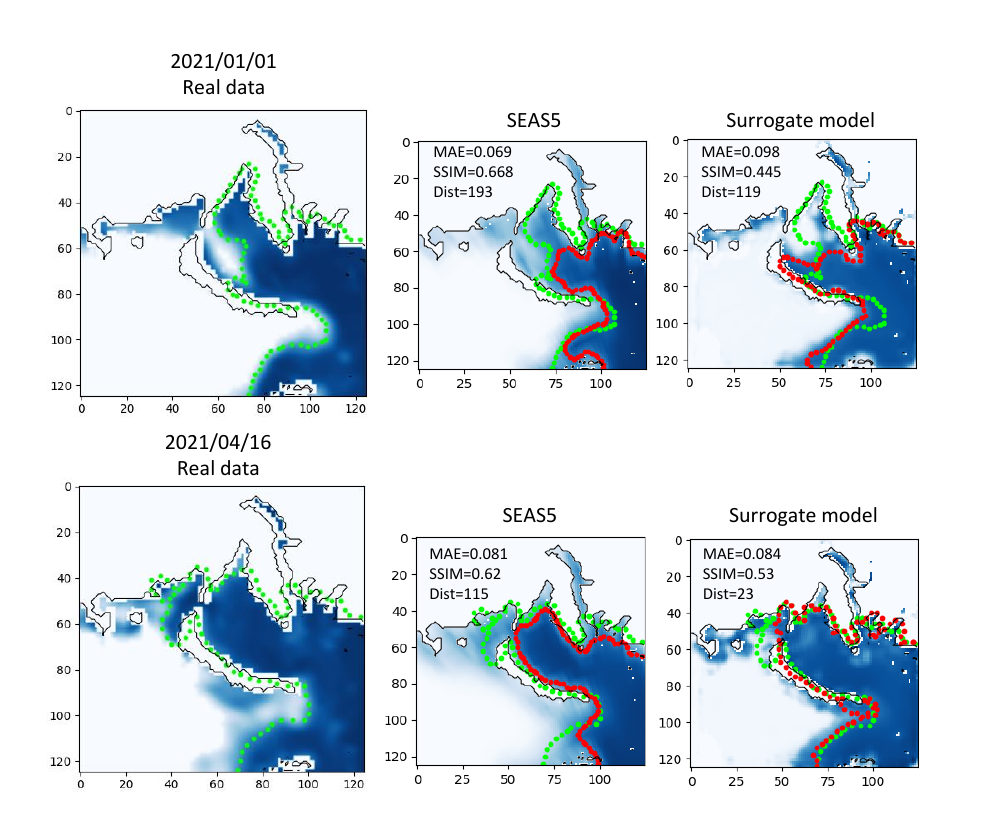}}
    \caption{Visualization of the identified ice edge on real data (green points) in comparison with forecasts of SEAS5 and a surrogate model (red points)}
    \label{fig:schema}
\end{figure*}

Figure~\ref{fig:schema} shows an example of the result of edge identification as 100 points with the distance metric calculation.

\begin{figure}[!h]
    \centering{\includegraphics[scale=0.49]{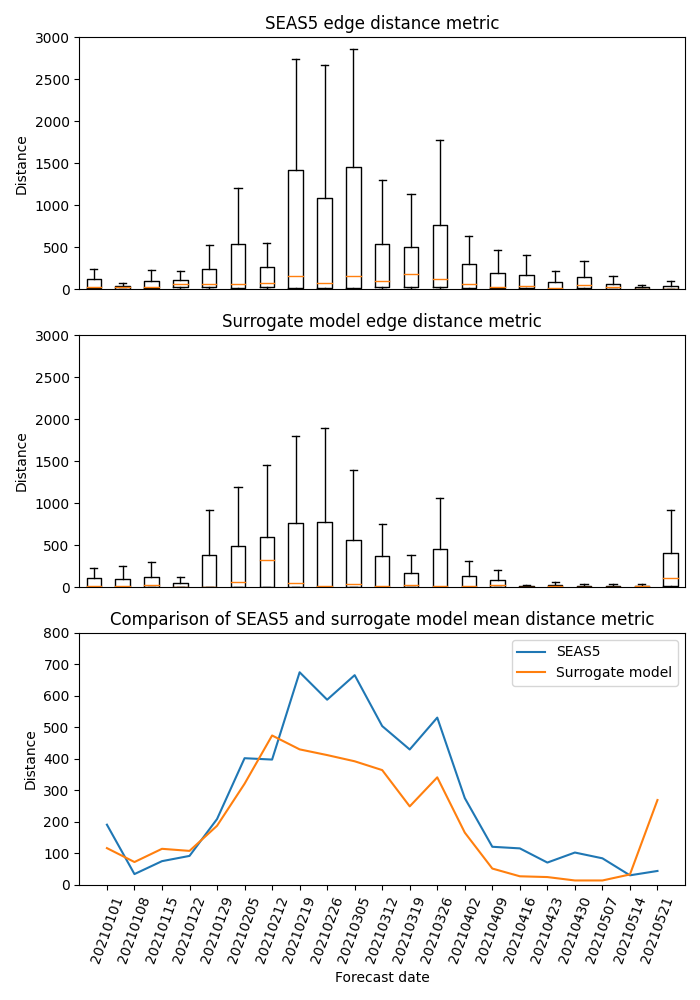}}
    \caption{Boxplots of distance criteria for each point of ice edge - comparison of SEAS5 and surrogate model prediction}
    \label{fig:schema_box}
\end{figure}

The distance values were displayed as boxplots for each prediction step to assess the range of agreement between the ice edge contour points. Figure~\ref{fig:schema_box} provides a visualization for the 2021 ice melt period.

The \textit{LANE-SI}-designed surrogate model provides a prediction close to SEAS5 by the average value of edge distance criteria. Also, the edge distance criteria variance for each point is lower for the proposed surrogate model.

\section{Conclusion}
\label{sec_concl}

This paper proposes an approach to predictive modeling of sea ice concentration named \textit{LANE-SI}. It is based on an ensemble of convolutional neural networks with different loss functions and naive forecast (that represent the average over the last several years).

Quality assessment was conducted using several objectives: (1) the coincidence of the absolute values of the parameter with actual data (MAE), (2) the coincidence of the spatial distribution of the parameter (SSIM), and (3) the distance between ice edges The results of experiments for Kara sea confirms that, according to the SSIM metric, an ensemble always gives better quality than single models; according to the MAE metric, in most cases, an ensemble always gives better quality. The comparison results for the average distance between actual and predicted ice edges indicate that, on average, over the contour and taking into account the scatter of the metric for each contour point, the surrogate model based on a convolutional network performs better in predicting the ice edge. Also, the developed model was compared with a physics-based forecast SEAS5. Experiments have shown that the \textit{LANE-SI} approach allows forecasting with a quality close to the physics-based SOTA model. While we used the Kara Sea for validation, \textit{LANE-SI} can be used to design the forecasting model with a specified resolution and forecast horizon for any water area in the Arctic.

The future research development will be aimed at applying automated machine learning to automate the design of surrogate ensemble models for new conditions. Also, we will consider the applicability of \textit{LANE-SI} to other environmental processes (e.g., weather forecasting).

\bibliography{aaai24}

\end{document}